\documentclass{article}
\RequirePackage{rotating}
\usepackage[ruled]{algorithm2e}
\usepackage{float}
\usepackage{natbib}
\usepackage{graphicx,epstopdf}
\usepackage{bbm}
\usepackage{theorem}
\usepackage{amsfonts}
\usepackage{booktabs}
\usepackage{multirow}
\usepackage{siunitx}
\usepackage{mathrsfs}
\usepackage{fancyhdr}
\usepackage{amsmath}
\usepackage{flexisym}
\usepackage{dsfont}
\usepackage{amssymb}
\usepackage{theorem}
\usepackage{graphicx}
\usepackage[dvips]{epsfig}
\usepackage{latexsym}
\usepackage{exscale}
\usepackage[latin1]{inputenc}
\usepackage{csquotes}
\usepackage{color}
\usepackage{rotating}
\usepackage{hyperref}

\numberwithin{equation}{section}

\DeclareMathOperator*{\argmax}{argmax}
\DeclareMathOperator*{\argmin}{argmin}

\newtheorem{theorem}{Theorem}[section]

\newtheorem{definition}[theorem]{Definition}

\begin{document}

\title{Some Developments in Clustering Analysis on Stochastic Processes}

\author{Qidi Peng\footnote{Institute of Mathematical Sciences, Claremont Graduate University, Claremont, CA 91711. Email: qidi.peng@cgu.edu.} \and Nan Rao\footnote{School of Mathematical Sciences, Shanghai Jiao Tong University, Shanghai, China. Email: nan.rao@sjtu.edu.cn.} \and Ran Zhao\footnote{Institute of Mathematical Sceinces and Drucker School of Management, Claremont Graduate University, Claremont, CA 91711. Email: ran.zhao@cgu.edu.}}

\date{}

\maketitle

\begin{abstract}
\noindent We review some developments on clustering stochastic processes and come with the conclusion that asymptotically consistent clustering algorithms can be obtained when the processes are ergodic and the dissimilarity measure satisfies the triangle inequality. Examples are provided when the processes are distribution ergodic, covariance ergodic and locally asymptotically self-similar, respectively.

\begin{flushleft}
\textbf{Keywords: } stochastic process, unsupervised clustering, stationary ergodic processes, local asymptotic self-similarity

\end{flushleft}
\end{abstract}







\section{Introduction}
\label{introduction}
A stochastic process is an infinite  sequence of random variables indexed by \enquote{time}. The time indexes can be either discrete or continuous.  Stochastic process type data have been broadly explored in biological and medical research \citep{Damian2007,Zhao2014,Jaaskinen2014,McDowell2018}. Unsupervised learning on stochastic processes (or time series) has increasingly attracted people from various areas of research and practice.  Among the above unsupervised learning problems, one subject, called cluster analysis, aims to discover patterns of stochastic process type data. There is a rich literature in bioinformatics, biostatistics and genetics on clustering stochastic process type data. We refer the readers to the review of  Aghabozorgi et al. \citep{Aghabozorgi2015} for updates of cluster analysis on stochastic processes til 2015. Recently Khaleghi et al. \citep{khaleghi2016,khaleghi2014,khaleghi2012a} obtained asymptotically consistent algorithms for clustering distribution stationary ergodic processes, in the case where the correct number of clusters is known or unknown. In their framework, the key idea is to define a proper dissimilarity measure $d$ between any 2 observed  processes, which characterizes the features of stationarity and ergodicity. Further Peng et al. \citep{Peng2018-1,Peng2018-2} derived consistent algorithms for clustering covariance stationary ergodic processes and locally asymptotically self-similar processes.

In this framework we review the recent developments in cluster analysis on the following 3 types of stochastic processes: \begin{description}
\item[Type (1)] distribution stationary ergodic processes;
\item[Type (2)] covariance stationary ergodic processes;
\item[Type (3)] locally asymptotically self-similar processes.
\end{description}
According to the nature of each type of processes, the ground-truths in the 3 clustering problems are defined differently.
In the ground-truth of Type $(1)$, two processes of identical process distributions are in one cluster; in the ground-truth of Type $(2)$, two processes having the same means and covariance structures are in one cluster; for the third type, the pattern is the means and covariance structures of the tangent processes.

From the summary we conclude that a sufficient condition for the clustering algorithms (provided below) being consistent, is that the corresponding dissimilarity measure and its estimates satisfy the triangle inequality and its estimator are consistent (they converge to the theoretical dissimilarity as the path length goes up to the infinity).

\section{Asymptotically Consistent Algorithms}

In  \citep{khaleghi2016}, assuming the correct number of clusters $\kappa$ is known, two types of datasets are considered in the cluster analysis: offline dataset and online dataset. In the offline dataset, the number of sample paths and the length of each sample path do not vary with respect to time. However in the online dataset, both can vary.  In  \citep{khaleghi2016} for each type of datasets, by using a particular dissimilarity measure, asymptotically consistent algorithms (Algorithm 1 for offline dataset and Algorithm 2 for online dataset) have been derived, aiming to cluster distribution stationary ergodic processes. Here asymptotic consistency means the output clusters from the algorithm converge to the ground-truths either in probability (weak sense) or almost surely (strong sense). Based on Khaleghi et al.'s approaches, Peng et al. \citep{Peng2018-1,Peng2018-2} suggested asymptotically consistent algorithms that are valid for a more general class of processes and dissimilarity measures.

Let $X_1,X_2$ be one of the 3 types of processes in the above section.  We denote by $d(X_1,X_2)$ a dissimilarity measure between 2 stochastic processes $X_1,X_2$, which satisfies the triangle inequality. And we denote by $\widehat d(\mathbf x_1,\mathbf x_2)$ the estimate of $d(X_1,X_2)$, where for $i=1,2$, $\mathbf x_i=(x_1^{(i)},\ldots,x_{n_i}^{(i)})$ is an observed sample path of the process $X_i$, with length $n_i$. Moreover, assume that $\widehat d$ also verifies the triangle inequality and it is consistent: for all $\mathbf x_1,\mathbf x_2$, sampled from $X_1,X_2$ respectively,
$$
\widehat d(\mathbf x_1,\mathbf x_2)\xrightarrow[\min\{n_1,n_2\}\to\infty]{\mathbb P~\mbox{or}~a.s.}d(X_1,X_2),
$$
where $\xrightarrow[]{\mathbb P}$ and $\xrightarrow[]{a.s.}$ denote the convergence in probability and almost sure convergence, respectively.

The clustering algorithms suggested by Peng et al. \citep{Peng2018-1,Peng2018-2} are given below.

\begin{algorithm}[h]\caption{Offline clustering} \label{algo::offline_known_k}

\LinesNumbered
\KwIn{\textbf{\textit{sample paths}} $S= \left \{ \mathbf{z}_1,\ldots,\mathbf{z}_N \right \}$; \textbf{\textit{number of clusters}} $\kappa$.}

$(c_1,c_2) \longleftarrow \argmax\limits_{(i,j)\in\{1,\ldots,N\}^2, i<j}\widehat{d}(\mathbf z_i,\mathbf z_j)$\;
$C_1 \longleftarrow \left \{ c_1 \right \}$; $C_2\longleftarrow\{c_2\}$\;
\For{$k = 3,\ldots,\kappa$}{
$c_k \longleftarrow \displaystyle\argmax_{i=1,\ldots,N}\displaystyle\min_{j = 1,\ldots,k-1} \widehat{d}(\mathbf{z}_i, \mathbf{z}_{c_j})$
}
\textbf{\textit{Assign the remaining points to the nearest centers}:}

\For{$i = 1,\ldots,N$}{
$k \longleftarrow \argmin\limits_{k\in\{1,\ldots,\kappa\}}\left\{\widehat{d}(\mathbf{z}_i, \mathbf{z}_j):~j \in C_k\right\}$;\\
$C_k \longleftarrow C_k \cup \left \{ i\right \}$
}
\KwOut{The $\kappa$ clusters $f(S,\kappa,\widehat{d})=\{C_1,C_2, \ldots, C_\kappa\}$.}
\end{algorithm}

\begin{algorithm}[h]
\caption{Online clustering}
\label{algo::online_known_k}
\LinesNumbered
\KwIn{\textbf{\textit{Sample paths}} $\Big\{S(t)=\{\mathbf z_1^t,\ldots,\mathbf z_{N(t)}^t\}\Big\}_t$; \textbf{\textit{number of clusters}} $\kappa$.}

\For{$t = 1,\ldots,\infty$}{
\textbf{\textit{Obtain new paths:} $S(t) \longleftarrow \Big\{\mathbf{z}_1^t,\dots,\mathbf{z}_{N(t)}^t\Big \}$}\;
\textbf{\textit{Initialize the normalization factor}: $\eta \longleftarrow 0$}\;
\textbf{\textit{Initialize the final clusters}: $C_k(t) \longleftarrow \emptyset,~k = 1,\ldots,\kappa$}\;
\textbf{\textit{Generate }$N(t) -\kappa + 1$ \textit{candidate cluster centers}:}

\For{$j = \kappa,\ldots,N(t)$}{
$\big \{C_1^j,\ldots, C_{\kappa}^j\big \} \longleftarrow\mbox{\textbf{Alg1}}\big( \big \{\mathbf{z}_1^t,\ldots, \mathbf{z}_j^t \big \}, \kappa \big)$\;
$c_k^j \longleftarrow \min \big \{ i\in C_k^j \big \}, k = 1,\ldots,\kappa$\;
$\gamma_j \longleftarrow \min\limits_{k,k'\in \{1,\ldots,\kappa\},k\neq k'} \widehat{d}\big(\mathbf{z}_{c_k^j}^t,\mathbf{z}_{c_{k'}^j}^t\big)$\;
$w_j \longleftarrow 1/j(j+1)$\;
$\eta \longleftarrow \eta+w_j\gamma_j$}
\textbf{\textit{Assign each point to a cluster}:}

\For{$i = 1,\ldots,N(t)$}{
$k \longleftarrow \argmin\limits_{k'\in \{1,\ldots,\kappa\}}\frac{1}{\eta} \sum\limits_{j=\kappa}^{N(t)}w_j \gamma_j \widehat{d}\big(\mathbf{z}_{i}^t,\mathbf{z}_{c_{k'}^j}^t\big)$\;
$C_k(t) \longleftarrow C_k(t)\cup \left\{ i \right\}$
}}
\KwOut{ \textit{The} $\kappa$ \textit{clusters} $f(S(t),\kappa,\widehat{d})=\left \{C_1(t), \dots, C_{\kappa}(t) \right \}$, $t=1,2,\ldots,\infty$.}
\end{algorithm}

\begin{theorem}
\label{thm:offline1}
Algorithms \ref{algo::offline_known_k} and \ref{algo::online_known_k} are asymptotically consistent for the processes of Types $(1)$ and $(2)$ respectively, provided that the correct number $\kappa$ of clusters is known, and the sample dissimilarity measure $\widehat d$ is consistent and both $\widehat d$ and $d$ satisfy the triangle inequality.
\end{theorem}
\textbf{Proof.} The consistency of Algorithms \ref{algo::offline_known_k} and \ref{algo::online_known_k} applied for clustering processes of Type $(1)$ is proved in \citep{khaleghi2016}; the consistency of the two algorithms applied for clustering processes of Type $(2)$ in proved in \citep{Peng2018-1}. $\square$

It is worth noting that in the above proof, the key features used are the fact that both $d$ and $\widehat d$ verify the triangle inequality and $\widehat d$ is a consistent estimator of $d$.

 For clustering the processes of Type $(3)$, an additional assumption is needed, which will be  introduced in Section \ref{local_self_similar}.

 For clustering the processes of Type $(1)$, the specific form of $d$ and $\widehat d$ are given in \citep{khaleghi2016}. Then we mainly introduce the other 2 pairs of $(d,
 \widehat d)$ for clustering analysis on the processes of Types $(2)$ and $(3)$.

\section{Dissimilarity Measure for Covariance Stationary Ergodic Processes}
\label{d:covariance}
The definition of covariance stationary ergodic process is given below.
\begin{definition}
 A stochastic process $\{X_t\}_{t\in\mathbb N}$ is covariance stationary ergodic if:
 \begin{itemize}
     \item its mean and covariance are invariant subject to any time shift;
     \item any of its sample autocovariance converges in probability to the theoretical autocovariance function as the sample length goes to $+\infty$.
 \end{itemize}
\end{definition}
The dissimilarity measure $d$ and its sample estimate $\widehat d$ suggested in Peng et al. \citep{Peng2018-1} to measure the distance between 2 covariance stationary ergodic processes are given below:
\begin{definition}
\label{definition1}
The dissimilarity measure $d$ between a pair of covariance stationary ergodic processes $X^{(1)}$, $X^{(2)}$ is defined as follows:
\begin{eqnarray*}
\label{def:d}
&&d(X^{(1)},X^{(2)})\nonumber\\
&&:= \displaystyle\sum_{m,l = 1}^{\infty} w_m w_l \rho\left(Cov(X_l^{(1)},\ldots,X_{l+m-1}^{(1)}),Cov(X_l^{(2)},\ldots,X_{l+m-1}^{(2)})\right),
\end{eqnarray*}
where:
\begin{itemize}
\item The sequence of positive weights $\{w_j\}$ is chosen such that  $d(X^{(1)},X^{(2)})$ is finite.
\item The distance $\rho$ between 2 equal-sized covariance matrices $M_1,M_2$ is defined to be $
\rho(M_1,M_2):=\|M_1-M_2\|_F$, with $\|\cdot\|_F$ being the \textit{Frobenius norm}.
\end{itemize}
\end{definition}

\begin{definition}
For two processes' paths $\mathbf x_j=(X_1^{(j)},\ldots,X_{n_j}^{(j)})$ for $j=1,2$, let $n=\min\{n_1,n_2\}$, then the empirical covariance-based dissimilarity measure between $\mathbf x_1$ and $\mathbf x_2$ is given by
\begin{equation*}
\label{dxx}
\widehat{d}(\mathbf x_{1},\mathbf x_{2}):=\sum_{m= 1}^{m_n} \sum_{l= 1}^{n-m+1} w_m w_l \rho\left (\nu(X^{(1)}_{l\ldots n},m), \nu(X^{(2)}_{l\ldots n},m)\right),
\end{equation*}
where:
\begin{itemize}
 \item $m_n$, chosen to be $o(n)$, denotes the size of covariance matrices considered in the estimator.
 \item $ \nu(\mathbf x,l,m):=\frac{\sum_{i=l}^{n-m+1}(X_i\ldots X_{i+m-1})^T(X_i\ldots X_{i+m-1})}{n-m-l+2}$ are the estimators of stationary covariance matrices.
 \end{itemize}
\end{definition}

\section{Dissimilarity Measure for Locally Asymptotically Self-similar Processes}
\label{local_self_similar}
In this section we review the work on clustering processes of  Type $(3)$.
Locally asymptotically self-similar processes  play a key role in the study of fractal geometry and wavelet analysis. They are generally not covariance stationary, however, one can still apply the dissimilarity measure $d$ introduced in Section \ref{d:covariance} under some assumption (see \citep{Peng2018-2}).

The following definition of locally asymptotically self-similar process is given in \citep{Boufoussi2008}.
\begin{definition}
\label{locally_asymptotically_self_similar}
A continuous-time stochastic process $\left\{Z_t^{(H(t))}\right\}_{t\ge0}$ with its index $H(\cdot)$ being a continuous function valued in $(0,1)$, is called locally asymptotically self-similar, if for each $t\ge0$, there exists a non-degenerate self-similar process $\left\{Y_u^{(H(t))}\right\}_{u\ge0}$ with self-similarity index $H(t)$,  such that
\begin{equation}
\label{local_self}
\left\{\frac{Z_{t+\tau u}^{(H(t+\tau u))}-Z_t^{(H(t))}}{\tau^{H(t)}}\right\}_{u\ge0}\xrightarrow[\tau\to0^+]{\mbox{f.d.d.}}\left\{Y_u^{(H(t))}\right\}_{u\ge0},
\end{equation}
where the convergence $\xrightarrow[]{\mbox{\textit{f.d.d.}}}$ is in the sense of all the finite dimensional distributions.
\end{definition}
Here $\{Y_u^{(H(t))}\}_u$ is called the \textit{tangent process} of $\{Z_t^{(H(t))}\}_t$ at $t$ (see \citep{Falconer2002}). Throughout \citep{Peng2018-2} it is assumed that the datasets are sampled from a known number of processes satisfying the following condition:
\newline
\noindent\textbf{Assumption $(\mathcal A)$:} The processes are locally asymptotically self-similar with distinct functional indexes $H(\cdot)$; their tangent processes' increment processes are \textit{covariance stationary ergodic}.

Then from (\ref{local_self}), Peng et al. \citep{Peng2018-2} showed the following: under  Assumption ($\mathcal A$), for $\tau$ being sufficiently small,
\begin{equation}
\label{local_self_2}
\left\{Z_{t+ \tau(u+h)}^{(H(t+\tau(u+h)))}-Z_{t+ \tau u}^{(H(t+\tau u))}\right\}_{u\in[0,Kh]}\stackrel{\mbox{f.d.d.}}{\approx}\left\{\tau^{H(t)}X_u^{(H(t))}\right\}_{u\in[0,Kh]},
\end{equation}
where $K$ is an arbitrary positive integer. Statistically, (\ref{local_self_2}) can be interpreted as: given a discrete-time path $Z_{t_1}^{(H(t_1))},\ldots,Z_{t_n}^{(H(t_n))}$ with $t_i=ih\Delta t$ for each $i\in\{1,\ldots,n\}$, sampled from a locally asymptotically self-similar process $\{Z_t^{(H(t))}\}$, its localized increment path with time index around $t_i$, i.e.,
\begin{equation}
\label{increment:z}
\mathbf z^{(i)}:=\left(Z_{t_{i+1}}^{(H(t_{i+1}))}-Z_{t_i}^{(H(t_i))},\ldots,Z_{t_{i+1+K}}^{(H(t_{i+1+K}))}-Z_{t_{i+K}}^{(H(t_{i+K}))}\right),
\end{equation}
is \textit{approximately} distributed as a covariance stationary ergodic increment process of the self-similar process $\left\{{\Delta t}^{H(t_i)}X_u^{(H(t_i))}\right\}_{u\in[0,Kh]}$. This fact inspires one to define the sample dissimilarity measure between two paths of locally asymptotically self-similar processes $\mathbf z_1$ and $\mathbf z_2$ as below:
\begin{equation}
\label{dissimilairty_local}
\widehat{d^{*}}(\mathbf z_1,\mathbf z_2):=\frac{1}{n-K-1}\sum_{i=1}^{n-K-1}\widehat{d}(\mathbf z_1^{(i)},\mathbf z_2^{(i)}),
\end{equation}
where $\mathbf z_1^{(i)}$, $\mathbf z_2^{(i)}$ are the localized increment paths defined as in (\ref{increment:z}).

Accordingly, the consistency of Algorithms \ref{algo::offline_known_k} and \ref{algo::online_known_k} can be expressed in the following way:
\begin{theorem}
\label{thm:offline2}
Under Assumption $(\mathcal A)$, Algorithms \ref{algo::offline_known_k} and \ref{algo::online_known_k} are approximately asymptotically consistent, if $\widehat{d}$ is replaced with $\widehat{d^{*}}$.
\end{theorem}
In Theorem \ref{thm:offline2}, \enquote{approximately} is in the sense of Eq. (\ref{local_self_2}).
\section{Simulation Study}
In the frameworks of khaleghi et al. \citep{khaleghi2016}, Peng et al. \citep{Peng2018-1} and Peng et al. \citep{Peng2018-2}, simulation study are provided. In \citep{khaleghi2016}, a distribution stationary ergodic process is simulated based on random walk; in \citep{Peng2018-1} the increment process of fractional Brownian motion \citep{Ayache2004} is picked as an example of covariance stationary ergodic process; in  \citep{Peng2018-2}, simulation study is performed on the so-called multifractional Brownian motion \citep{Peltier1995}, which is an excellent example of locally asymptotically self-similar process. The simulation study results for clustering distribution stationary ergodic processes are given in \citep{khaleghi2016}. Here we summarize the results for clustering the processes of Types $(2)$ and $(3)$, from Peng et al. \citep{Peng2018-1} and \citep{Peng2018-2} respectively.
\subsection{Clustering Processes of Type $(2)$: Fractional Brownian Motion}
Fractional Brownian motion (fBm) $\{B^H(s)\}_{s\ge0}$, as a natural extension of the Brownian motion, was first introduced by Kolmogorov in 1940 and then popularized by Mandelbrot and Taqqu \citep{Mandelbrot1968,Taqqu2013}. The influences made by the fractional Brownian motion model have been on a great many fields such as biological science, physical sciences and economics (see \citep{Hofling2013}). As a stationary increment process, it is shown that the increment process of the fBm is covariance stationary ergodic (see \cite{Maruyama1970,Slezak2017}).

In \citep{Peng2018-1}, the clustering algorithms are performed on a dataset of $100$ paths of fBms with $\kappa=5$ clusters. In the sample dissimilarity measure the so-called $\log^*$-transformation is applied to increase the efficiency of the algorithms. One considers the mis-clustering rates to be the measure of fitting errors.
The top figure in Figure \ref{fig::empirical_results} presents the comparison results of the offline and online algorithms,  based on the behavior of mis-clustering rates as time increases. Both algorithms show to be asymptotically consistent as the mis-clustering rates decrease.

\subsection{Clustering Processes of Type $(3)$: Multifractional Brownian Motion}
Multifractional Brownian motion (mBm) $\{W_{H(t)}(t)\}_{t\ge0}$, as a natural generalization of the fBm, was introduced in \citep{Peltier1995, ACLV00}. Then it was quickly applied to describe phenomena in for instance molecular biology \citep{Marquez2012}, biomedical engineering \citep{Buard2010} and biophysics \citep{Humeau2007}.

It can be obtained from \cite{Boufoussi2008} that the mBm is locally asymptotically self-similar satisfying Assumption $(\mathcal A)$.

\begin{figure}[h]
\centering
\includegraphics[scale = 0.2]{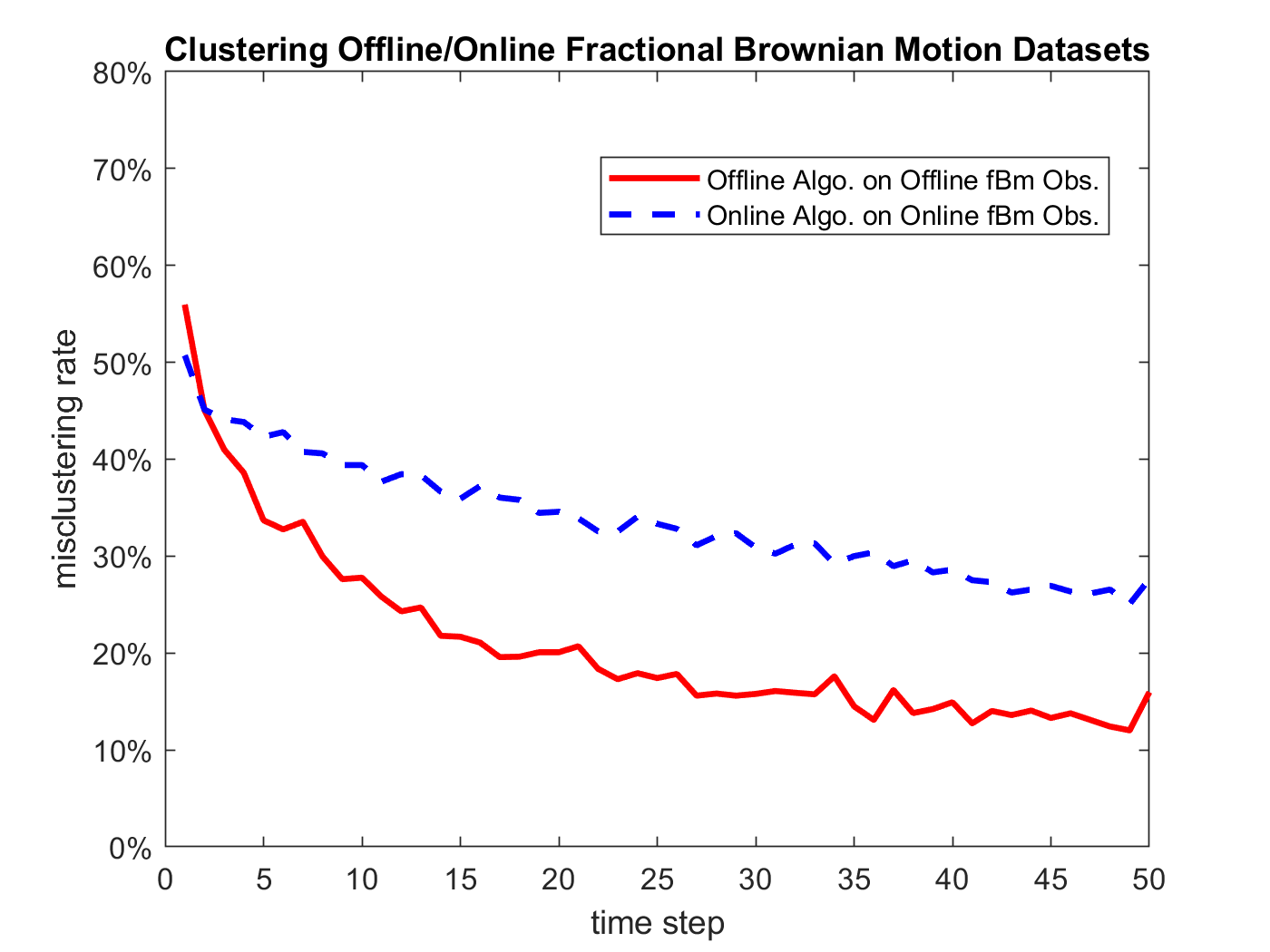}
\includegraphics[scale = 0.2]{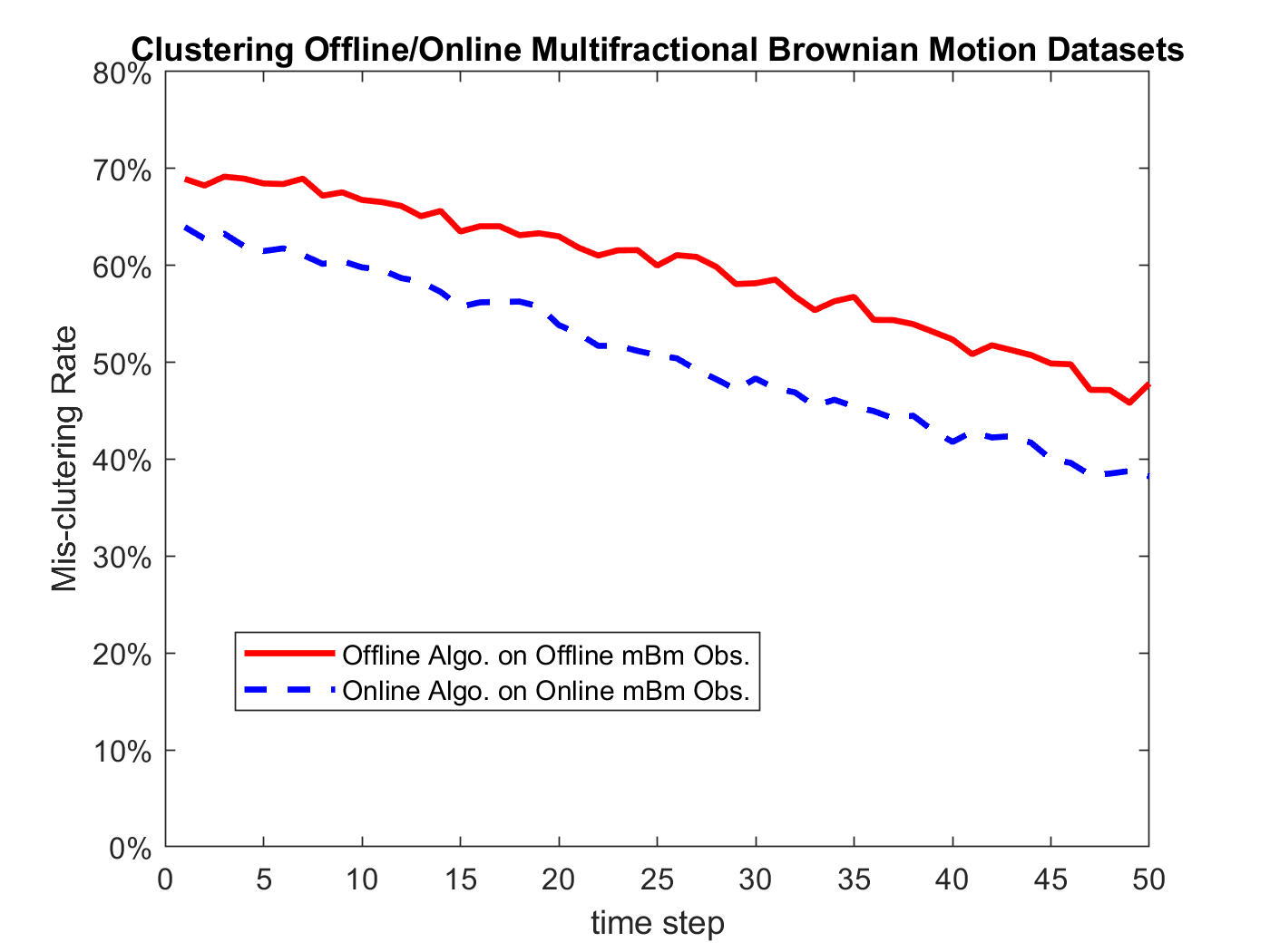}
\caption{The top graph illustrates the mis-clustering rates of offline algorithm (Algorithm \ref{algo::offline_known_k}, solid red line), and online algorithm (Algorithm \ref{algo::online_known_k}, dashed blue line) for fBm datasets. The bottom graph illustrates the mis-clustering rates of offline algorithm (Algorithm \ref{algo::offline_known_k}, solid red line), and online algorithm (Algorithm \ref{algo::online_known_k}, dashed blue line) for mBm datasets.}
\label{fig::empirical_results}
\end{figure}

The datasets of mBms for testing the 2 clustering algorithms are similar to those of fBms. The performance of the algorithms are shown in the bottom figure in Figure \ref{fig::empirical_results}. Similar conclusion can be drawn that both offline and online algorithms are approximately asymptotically consistent.

\newpage

\bibliographystyle{apalike}
\bibliography{references}

\end{document}